\documentclass[conference]{IEEEtran}
\IEEEoverridecommandlockouts
\usepackage{cite}
\usepackage{amsmath,amssymb,amsfonts}
\usepackage{algorithmic}
\usepackage{graphicx}
\usepackage{textcomp}
\usepackage{xcolor}
\usepackage{placeins}
\usepackage{comment}
\usepackage{hyperref}
\def\BibTeX{{\rm B\kern-.05em{\sc i\kern-.025em b}\kern-.08em
    T\kern-.1667em\lower.7ex\hbox{E}\kern-.125emX}}

\begin{document}

\title{Beyond Labels: Advancing Cluster Analysis with the Entropy of Distance Distribution (EDD)\\
\thanks{DFG: Deutsche Forschungsgemeinschaft}
}

%\begin{comment}
\author{\IEEEauthorblockN{1\textsuperscript{st} Claus Metzner}
\IEEEauthorblockA{\textit{Neuroscience Lab} \\
\textit{ University Hospital Erlangen}\\
Erlangen, Germany \\
claus.metzner@fau.de}
\and
\IEEEauthorblockN{2\textsuperscript{nd} Achim Schilling}
\IEEEauthorblockA{\textit{Neuroscience Lab} \\
\textit{University Hospital Erlangen }\\
Erlangen, Germany \\
achim.schilling@fau.de}
\and
\IEEEauthorblockN{3\textsuperscript{rd} Patrick Krauss}
\IEEEauthorblockA{\textit{Neuroscience Lab} \\
\textit{ University Hospital Erlangen}\\
Erlangen, Germany \\
patrick.krauss@fau.de}}
%\end{comment}

\begin{comment}
\author{\IEEEauthorblockN{1\textsuperscript{st} Anonymous Author}
}
\end{comment}

\maketitle

%=========================================================================
\begin{abstract}
In the evolving landscape of data science, the accurate quantification of clustering in high-dimensional data sets remains a significant challenge, especially in the absence of predefined labels. This paper introduces a novel approach, the Entropy of Distance Distribution (EDD), which represents a paradigm shift in label-free clustering analysis. Traditional methods, reliant on discrete labels, often struggle to discern intricate cluster patterns in unlabeled data. EDD, however, leverages the characteristic differences in pairwise point-to-point distances to discern clustering tendencies, independent of data labeling.

Our method employs the Shannon information entropy to quantify the 'peakedness' or 'flatness' of distance distributions in a data set. This entropy measure, normalized against its maximum value, effectively distinguishes between strongly clustered data (indicated by pronounced peaks in distance distribution) and more homogeneous, non-clustered data sets. This label-free quantification is resilient against global translations and permutations of data points, and with an additional dimension-wise z-scoring, it becomes invariant to data set scaling.

We demonstrate the efficacy of EDD through a series of experiments involving two-dimensional data spaces with Gaussian cluster centers. Our findings reveal a monotonic increase in the EDD value with the widening of cluster widths, moving from well-separated to overlapping clusters. This behavior underscores the method's sensitivity and accuracy in detecting varying degrees of clustering. EDD's potential extends beyond conventional clustering analysis, offering a robust, scalable tool for unraveling complex data structures without reliance on pre-assigned labels.
\end{abstract}


\section*{Introduction}

The quest for understanding and interpreting complex data structures has long been at the forefront of data science and artificial intelligence research \cite{guan2022novel, gauss2023dcsi}. A pivotal step in this journey is the effective quantification of data clustering, particularly in high-dimensional spaces. Our previous work in this domain was marked by the introduction of the Generalized Discrimination Value (GDV), a novel metric designed to non-invasively measure the separability of different data classes within the layers of an artificial neural network \cite{schilling2021quantifying, krauss2021analysis, krauss2018analysis}.

The GDV, as detailed in our preceding research, serves as a powerful tool in analyzing neural network dynamics, particularly in the context of class separation during the training process. It revealed that, regardless of initial conditions, each layer of a neural network attains a highly reproducible GDV value at the end of the training period. This consistency was observed in multi-layer perceptrons trained with error backpropagation, where the classification of complex datasets was found to necessitate a temporal reduction in class separability. This phenomenon was characterized by a distinct 'energy barrier' in the initial phase of the GDV curve. Astonishingly, for a given dataset, the GDV follows a 'master curve', independent of the neural network's architectural complexity \cite{schilling2021quantifying}.

Our observations also underscored the GDV's invariance with respect to dimensionality, highlighting its potential as a comparative tool for examining the internal representational dynamics of diverse neural network architectures. This aspect of GDV is particularly significant for tasks such as neural architecture search, network compression, and even in drawing parallels with brain function models.

Building on the foundational insights gained from the GDV, our current study ventures into a new realm of clustering analysis. We aim to extend the principles of data separability and clustering quantification to scenarios where discrete labels are absent or infeasible to determine. This paper introduces the Entropy of Distance Distribution (EDD), a methodological leap in label-free quantification of clustering. EDD leverages the unique distributions of pairwise point-to-point distances within data sets, offering a robust and scalable tool for unraveling clustering patterns in high-dimensional, unlabeled data spaces. This advancement not only complements the GDV's capabilities but also significantly broadens the horizon for analyzing complex data structures in a variety of scientific and industrial applications.

%=========================================================================
\section*{Methods}

The Entropy of Distance Distribution (EDD) method presented in this study is formulated to quantify the degree of clustering in high-dimensional data sets without the need for discrete labels. This approach is rooted in analyzing the distributions of pairwise point-to-point distances within the data set.

\subsection*{Pairwise Distance Calculation}
Given a data set, we first compute the pairwise distances between all points. Let $x_i$ and $x_j$ represent two points in the data set. The distance between these points is denoted as $d_{ij}$, calculated using a suitable metric (e.g., Euclidean distance).

\subsection*{Distance Distribution and Histogram}
The next step involves constructing a histogram of all calculated distances $d_{ij}$. This histogram, denoted as $H(d)$, represents the frequency of occurrence of each distance value. It is normalized to create a binned probability distribution $p(d_k)$, where $d_k$ refers to the k-th bin in the histogram.

\subsection*{Entropy Calculation}
The core of the EDD method lies in quantifying the 'peakedness' or 'flatness' of the normalized distance distribution using Shannon information entropy. The entropy, $H(p)$, is calculated as follows:

\begin{equation}
H(p) = - \sum_k p(d_k) \log_2 p(d_k)
\end{equation}

This entropy value is small when the probability mass is concentrated within a few peaks (indicative of strong clustering) and is maximal for a uniform distribution (indicative of no clustering).

\subsection*{Normalization of Entropy}
To facilitate comparison across different data sets or conditions, the entropy is normalized by its maximum possible value. This maximum value is the logarithm of the number of histogram bins, $\log_2(N)$, where N is the total number of bins used in $H(d)$. The normalized EDD is then given by:

\begin{equation}
EDD = H(p) / \log_2(N)
\end{equation}

This normalized EDD value ranges between zero (indicating perfect clustering) and one (indicating no clustering).

\subsection*{Generalized Discrimination Value}

In order to compare the label-free EDD with a label-based clustering measure, we calculate the Generalized Discrimination Value (GDV) as published and explained in detail in \cite{schilling2021quantifying}. Briefly, we consider $N$ points $\mathbf{x_{n=1..N}}=(x_{n,1},\cdots,x_{n,D})$, distributed within $D$-dimensional space. A label $l_n$ assigns each point to one of $L$ distinct classes $C_{l=1..L}$. In order to become invariant against scaling and translation, each dimension is separately z-scored and, for later convenience, multiplied with $\frac{1}{2}$:
\begin{align}
s_{n,d}=\frac{1}{2}\cdot\frac{x_{n,d}-\mu_d}{\sigma_d}.
\end{align}
Here, $\mu_d=\frac{1}{N}\sum_{n=1}^{N}x_{n,d}\;$ denotes the mean, and $\sigma_d=\sqrt{\frac{1}{N}\sum_{n=1}^{N}(x_{n,d}-\mu_d)^2}$ the standard deviation of dimension $d$.
Based on the re-scaled data points $\mathbf{s_n}=(s_{n,1},\cdots,s_{n,D})$, we calculate the {\em mean intra-class distances} for each class $C_l$ 
\begin{align}
\bar{d}(C_l)=\frac{2}{N_l (N_l\!-\!1)}\sum_{i=1}^{N_l-1}\sum_{j=i+1}^{N_l}{d(\textbf{s}_{i}^{(l)},\textbf{s}_{j}^{(l)})},
\end{align}
and the {\em mean inter-class distances} for each pair of classes $C_l$ and $C_m$
\begin{align}
\bar{d}(C_l,C_m)=\frac{1}{N_l  N_m}\sum_{i=1}^{N_l}\sum_{j=1}^{N_m}{d(\textbf{s}_{i}^{(l)},\textbf{s}_{j}^{(m)})}.
\end{align}
Here, $N_k$ is the number of points in class $k$, and $\textbf{s}_{i}^{(k)}$ is the $i^{th}$ point of class $k$.
The quantity $d(\textbf{a},\textbf{b})$ is the euclidean distance between $\textbf{a}$ and $\textbf{b}$. Finally, the Generalized Discrimination Value (GDV) is calculated from the mean intra-class and inter-class distances  as follows:
\begin{align}
\mbox{GDV}=\frac{1}{\sqrt{D}}\left[\frac{1}{L}\sum_{l=1}^L{\bar{d}(C_l)}\;-\;\frac{2}{L(L\!-\!1)}\sum_{l=1}^{L-1}\sum_{m=l+1}^{L}\bar{d}(C_l,C_m)\right]
 \label{GDVEq}
\end{align}

\noindent whereas the factor $\frac{1}{\sqrt{D}}$ is introduced for dimensionality invariance of the GDV with $D$ as the number of dimensions.

\vspace{0.2cm}\noindent Note that the GDV is invariant with respect to a global scaling or shifting of the data (due to the z-scoring), and also invariant with respect to a permutation of the components in the $N$-dimensional data vectors (because the euclidean distance measure has this symmetry). The GDV is zero for completely overlapping, non-separated clusters, and it becomes more negative as the separation increases. A GDV of -1 signifies already a very strong separation (see \cite{schilling2021quantifying}).

%=========================================================================

\section*{Results}

To ensure the robustness of the EDD, we incorporate measures to make it invariant to certain transformations:

1) \textbf{Global Translations and Permutations:} Since EDD is based on distances between pairs of points, it naturally remains invariant to global translations and permutations of the data points.

2) \textbf{Scaling of the Data Set:} By performing dimension-wise z-scoring (subtracting the mean and dividing by the standard deviation for each dimension), we further enhance the EDD's invariance to scaling variations across different data sets.

Through these methodological steps, the EDD provides a comprehensive and label-free approach to quantify clustering in high-dimensional data spaces, extending beyond the capabilities of traditional metrics like the GDV.

\begin{figure*}[htb]
  \centering
  \includegraphics[width = 0.7\textwidth]{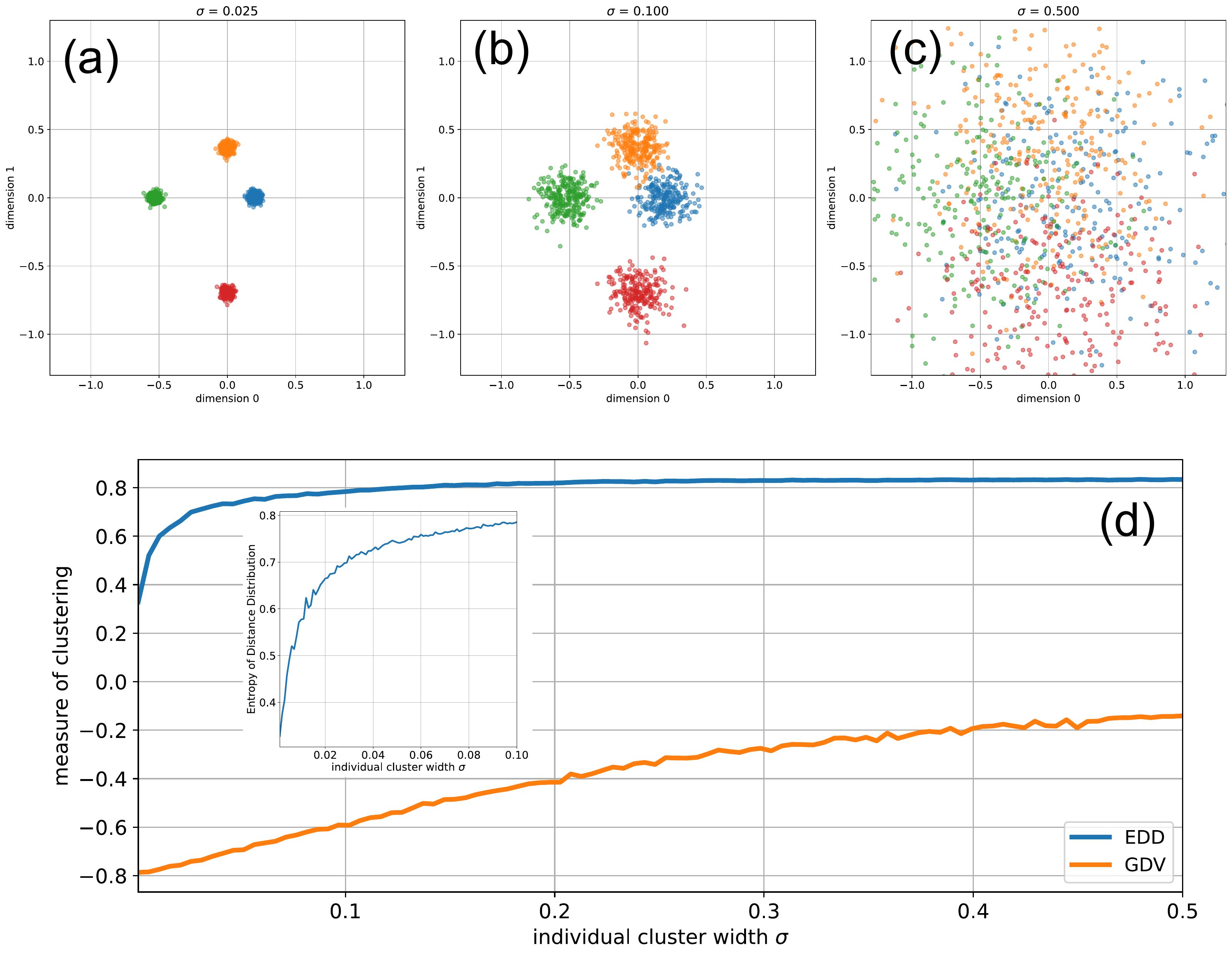}
  \label{fig1}
  \caption{\textbf{Quantification of clustering with (GDV) and without (EDD) labels.} \newline Four fixed Gaussian cluster centers in a two-dimensional data space. Starting from a regime where the clusters are still well-separated (a), the width of the clusters can be gradually increased until the clusters overlap (b,c). Both the GDV (orange) and the EDD (blue) increase monotonously as a function of the cluster width (d). The inset shows the EDD in the regime of small widths, where clusters are still well defined. Note that the EDD does not get any of the colored labels to work with.}
\end{figure*}

Figure \ref{fig1} shows a simple demonstration, in which the data set consists of three fixed Gaussian cluster centers in a two-dimensional data space. Starting from a regime where the clusters are well-separated, the width of the clusters can be gradually increased until the clusters overlap. We find that as a function of the cluster width, the EDD increases monotonously towards the upper limit of one.

\begin{figure*}[htb]
  \centering
  \includegraphics[width = 0.7\textwidth]{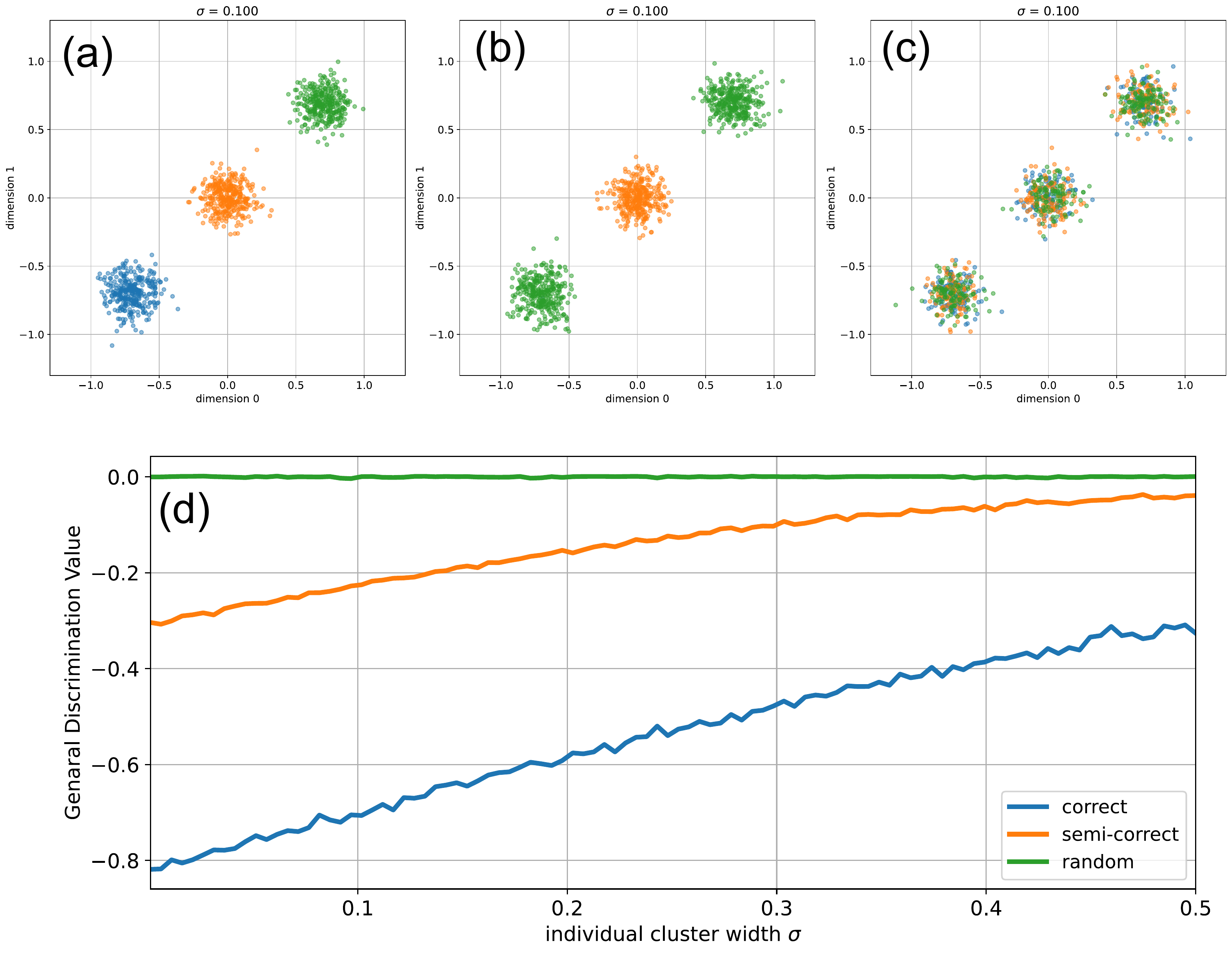}
  \label{fig2}
  \caption{\textbf{Dependence of the GDV on the provided labeling.}\newline Given are three clusters in a two-dimensional data space. We consider case (a) where all labels are assigned correctly, case (b) where two of the clusters have been incorrectly assigned the same labels, and case (c) where all labels are assigned randomly. For the correct labeling (d, blue curve), the GDV is generally low (signifying good clustering) and increases monotonously with the cluster width. For the semi-correct labeling (d, orange curve), the GDV also increases monotonously, but has considerably larger values (signifying a smaller degree of clustering). For the random case (d, green curve), the GDV has a constant value of zero, independent of the cluster width.}
\end{figure*}

%======================================================

\section*{Discussion}

The introduction of the Entropy of Distance Distribution (EDD) method marks a significant advancement in the field of data analysis, particularly in the context of high-dimensional and unlabeled datasets. This study demonstrates the EDD's ability to quantify clustering in a manner that is effective in extending the foundational concepts established by the Generalized Discrimination Value (GDV).

\subsection*{Key Findings}
Our results highlight the EDD's sensitivity and accuracy in detecting varying degrees of clustering. The method's reliance on the distribution of pairwise distances, and its subsequent quantification through Shannon information entropy (Quelle), provides a nuanced perspective of data structure that is not reliant on pre-assigned labels. This is a crucial development, as it addresses a common challenge in many real-world datasets where labeling is either impractical or impossible.

\subsection*{Comparison with GDV}
While the GDV offered valuable insights into class separability in neural networks, its applicability was limited to scenarios where labels were available and meaningful. The EDD transcends this limitation, offering a label-free approach. This is particularly useful in exploratory data analysis phases, where the inherent structure of the data is unknown or in cases where data is too complex for traditional labeling. 

\subsection*{Practical Implications}
The EDD's invariance to data set scaling and its resilience against global translations and permutations of data points make it a robust tool for a wide range of applications. From biological data analysis to social network analysis, the EDD can be employed to uncover underlying patterns that are not immediately apparent. One concrete application could be the investigation of the newly introduced Large Language Models (LLMs) based on the 'transformer' architecture \cite{vaswani2017attention} such as GPT4 or BERT \cite{bubeck2023sparks, devlin2018bert}. Up to now, these networks act as black boxes \cite{castelvecchi2016can} and it remains elusive, if e.g. these networks could be improved by a deeper understanding of the underlying principles or if we potentially could gain insights for language processing in the human brain from these LLMs, which would be in line with the philosophy of "cognitive computational neuroscience" \cite{kriegeskorte2018cognitive, schilling2023predictive}. Thus, clustering of the representations (neuron activation patterns) in the different layers of the network could be analyzed and quantified using the EDD. In a second step, the identified clusters can be investigated in detail to find the according label such as syntactic or semantic structures of the input as well as output data (cf. \cite{stoewer2022neural, stoewer2023neural, stoewer2023conceptual, surendra2023word}).
Especially the scalability of the EDD makes it suitable for large datasets such as big text corpora typical in the era of big data.

\subsection*{Limitations and Future Work}
While the EDD shows promising results, there are limitations to consider. The choice of distance metric and the number of histogram bins can influence the EDD's output, suggesting a need for further research to optimize these parameters. Additionally, the method's performance in extremely high-dimensional spaces or in datasets with intricate, overlapping clusters warrants further investigation.

\FloatBarrier
%=========================================================================
\vspace{-0.1cm}
\section*{Acknowledgments}
This work was funded by the Deutsche Forschungsgemeinschaft (DFG, German Research Foundation): grants KR\,5148/2-1 (project number 436456810), KR\,5148/3-1 (project number 510395418) and GRK\,2839 (project number 468527017) to PK, and grant SCHI\,1482/3-1 (project number 451810794) to AS.

%=========================================================================
\vspace{-0.1cm}
\section*{Author contributions}
All authors contributed equally to this manuscript.

%=========================================================================
\vspace{-0.15cm}
\section*{Competing interests}
The authors declare no competing financial interests.

\vspace{-0.15cm}
%=========================================================================
\bibliographystyle{unsrt}
\bibliography{literature}

\end{document}